\documentclass[letterpaper, 10 pt, preprint]{ieeeconf}  %

\IEEEoverridecommandlockouts                              %

\usepackage{graphics} %
\usepackage{epsfig} %
\usepackage{mathptmx} %
\usepackage{times} %
\usepackage{amsmath} %
\usepackage{amssymb}  %

\usepackage{listings}
\usepackage{float}
\usepackage{ragged2e}
\usepackage{subcaption}
\usepackage{placeins}
\usepackage{graphicx}
\usepackage{hyperref}
\usepackage{booktabs}
\usepackage{adjustbox} %
\usepackage{dblfloatfix} %
\usepackage{placeins}
\usepackage{xcolor}
\usepackage{cuted}
\usepackage{multirow}
\usepackage[T1]{fontenc}

\title{\LARGE \bf
Gradient-Driven 3D Segmentation and Affordance Transfer in Gaussian Splatting Using 2D Masks
}

\author{Joji Joseph$^{1}$, Bharadwaj Amrutur$^{2}$ and Shalabh Bhatnagar$^{3}$%
\thanks{$^{1}$Joji Joseph is with RBCCPS,
        Indian Institute of Science, Bangalore, India
        {\tt\small jojijoseph@iisc.ac.in}}%
\thanks{$^{2}$Bharadwaj Amrutur is with RBCCPS and ARTPARK, Indian Institute of Science, Bangalore, India
        {\tt\small amrutur@iisc.ac.in}}%
\thanks{$^{2}$Shalabh Bhatnagar is with RBCCPS and CSA, Indian Institute of Science, Bangalore, India
        {\tt\small shalabh@iisc.ac.in}}%
}

\begin{document}

\maketitle
\thispagestyle{empty}
\pagestyle{empty}

\begin{abstract}
3D Gaussian Splatting has emerged as a powerful 3D scene representation technique, capturing fine details with high efficiency. In this paper, we introduce a novel voting-based method that extends 2D segmentation models to 3D Gaussian splats. Our approach leverages masked gradients, where gradients are filtered by input 2D masks, and these gradients are used as votes to achieve accurate segmentation. As a byproduct, we discovered that inference-time gradients can also be used to prune Gaussians, resulting in up to 21\% compression. Additionally, we explore few-shot affordance transfer, allowing annotations from 2D images to be effectively transferred onto 3D Gaussian splats. The robust yet straightforward mathematical formulation underlying this approach makes it a highly effective tool for numerous downstream applications, such as augmented reality (AR), object editing, and robotics. The project code and additional resources are available at \url{https://jojijoseph.github.io/3dgs-segmentation}.

\end{abstract}

\section{Introduction}
3D Gaussian splatting (3DGS) is a popular technique for rendering novel viewpoints using 3D Gaussian distributions as rendering primitives\cite{kerbl3Dgaussians}. It is particularly known for its speed and flexibility in rendering. Beyond color information, 3DGS can be extended to learn 3D feature fields, enabling applications in 2D and 3D segmentation and localization tasks\cite{qin2023langsplat, zhou2024feature}.

However, while feature fields perform well in tasks like localization and 2D segmentation \cite{qin2023langsplat, zhou2024feature}, they often encounter challenges with clean 3D segmentation, resulting in artifacts such as floaters—small, disconnected fragments that degrade the quality of the 3D representation. Furthermore, training these models is computationally intensive and slow, limiting their practical use.

Our approach addresses these challenges by using 2D segmentation masks generated by 2D segmention models and applying inference-time gradient backpropagation. Inspired by 2D techniques for visualization, such as DeepDream\cite{deepdream} and Grad-CAM\cite{Selvaraju_2019}, our method identifies Gaussians responsible for specific 2D masks, enabling more accurate and efficient 3D segmentation.

Although we have utilized 2D segmentation models for mask generation in our experiments, it is also possible to leverage a feature field method like LangSplat \cite{qin2023langsplat} for mask generation. This would enable the conversion of feature-field-based methods, which are primarily suited for 2D segmentation, into effective tools for 3D segmentation.

In this paper, our key contributions are as follows.

\begin{enumerate}
    \item We present a novel method for lifting segmentation results from 2D models to 3D Gaussian splats.
    \item We incorporates 2D-to-3D affordance transfer, enhancing manipulation tasks in 3D environments.
    \item We demonstrate the use of inference-time gradients for pruning trained Gaussians.
\end{enumerate}

\section{Related Works}

Neural Radiance Fields (NeRF) \cite{mildenhall2020nerf} is a state-of-the-art method for novel view synthesis, where a neural network is trained on sparse input images to generate photorealistic views from unseen viewpoints. NeRF learns an implicit volumetric representation of a scene by encoding color and density at each 3D point. However, because NeRF represents scenes implicitly, reconfiguring or modifying objects within the scene is challenging and often requires retraining the network.

In contrast, 3D Gaussian Splatting \cite{kerbl3Dgaussians} offers an explicit 3D scene representation, where 3D Gaussians serve as the fundamental graphics primitives. Each Gaussian has associated attributes such as color, opacity, and orientation. Rearranging and editing objects in 3D Gaussian Splatting can be achieved by directly manipulating the Gaussians and their corresponding payloads. This explicit nature provides greater flexibility for interactive applications, such as object editing, augmented reality and real time reconfiguration.

A natural extension of radiance field rendering is feature field rendering, where additional feature embeddings are incorporated. Recent works like \cite{zhou2024feature, qin2023langsplat, shi2023language} have explored this idea by introducing feature fields with an additional embedding payload.

\begin{figure*}%
    \justifying
    \begin{subfigure}[b]{0.24\textwidth}
    \frame{\includegraphics[width=\textwidth, height=\textwidth]{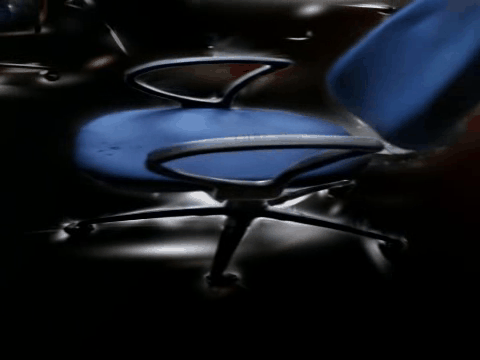}}
    \caption{Extraction - Feature 3DGS}
    \end{subfigure} %
    \begin{subfigure}[b]{0.24\textwidth}
    \frame{\includegraphics[width=\textwidth, height=\textwidth]{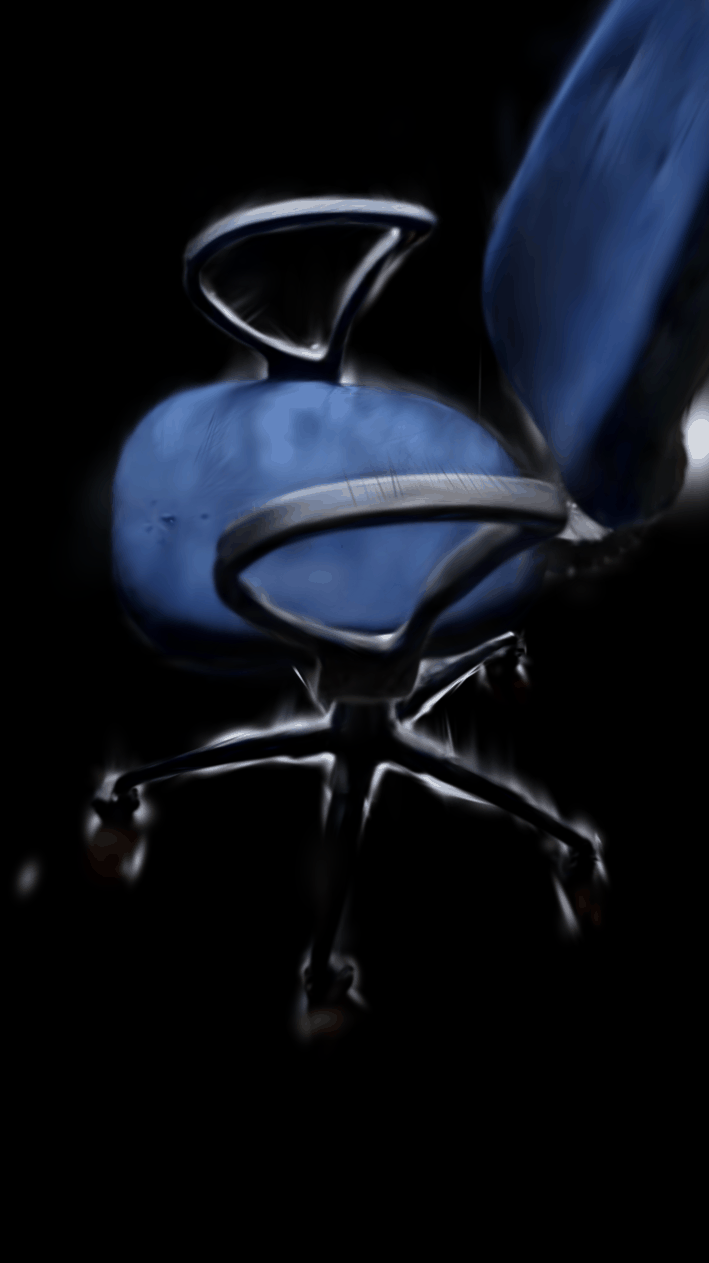}}
    \caption{Extraction - Ours}
    \end{subfigure}%
    \begin{subfigure}[b]{0.24\textwidth}
    \frame{\includegraphics[width=\textwidth, height=\textwidth]{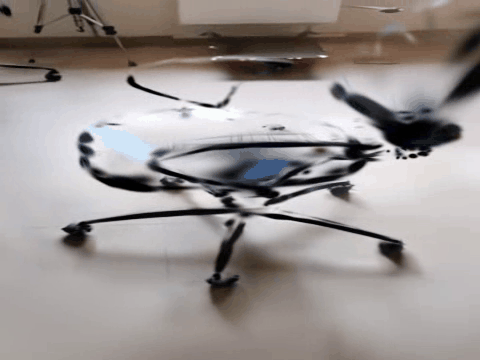}}
    \caption{Deletion - Feature 3DGS}
    \end{subfigure} %
    \begin{subfigure}[b]{0.24\textwidth}
    \frame{\includegraphics[width=\textwidth, height=\textwidth]{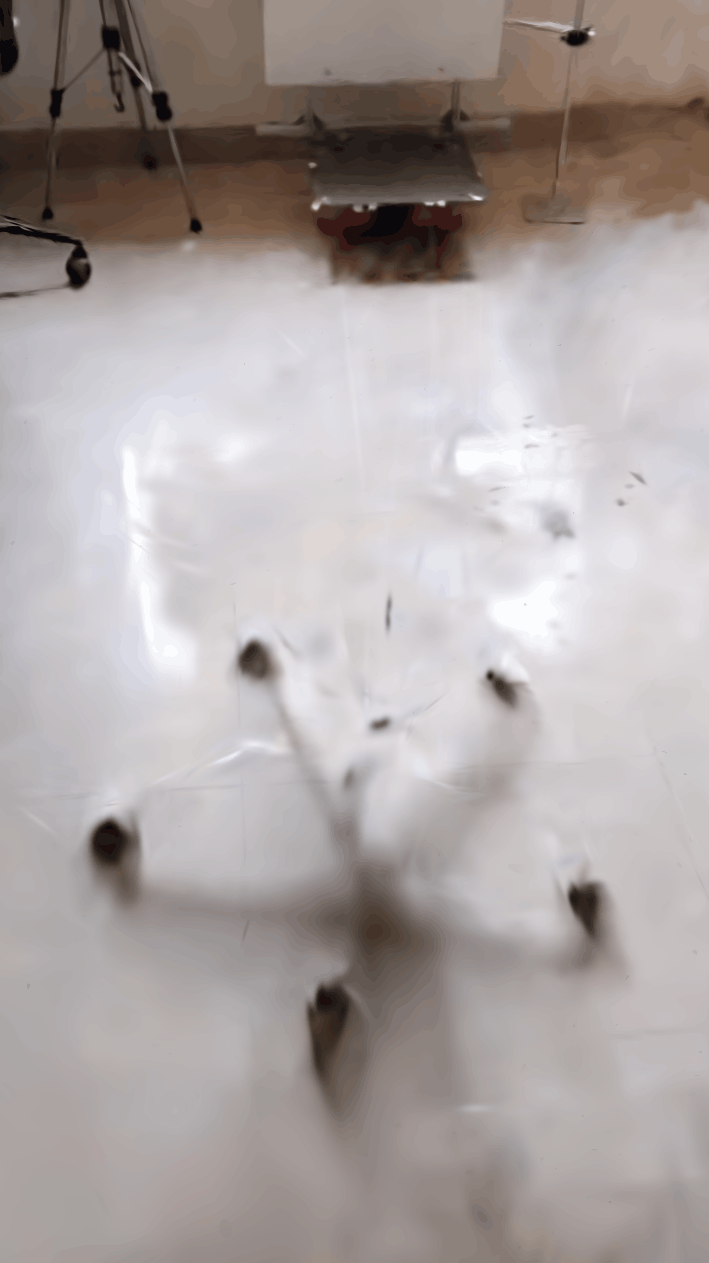}}
    \caption{Deletion - Ours}
    \end{subfigure}
    \caption{Comparison of our approach with Feature-3DGS. The segmentation produced by Feature-3DGS is less clean than our method because individual Gaussians may not have features that are fully representative of the final rendered features.}
    \label{fig:comparison_feature_3dgs_vs_ours}
\end{figure*}

For instance, Feature-3DGS \cite{zhou2024feature} focuses on training high-dimensional features, while LangSplat \cite{qin2023langsplat} emphasizes training compressed, low-dimensional features. Both approaches have shown excellent results for 2D segmentation of rendered outputs. However, they struggle with 3D object segmentation. As illustrated in Figure~\ref{fig:comparison_feature_3dgs_vs_ours}, we compare an object segmented using Feature-3DGS with our method.

Feature-3DGS performs 3D segmentation by matching language embeddings to the feature embeddings of the Gaussians. While effective for 2D segmentation, it often fails in 3D because the features of individual Gaussians do not fully correspond to the final rendered feature, which is a weighted sum of contributions from multiple Gaussians (see Equation~\ref{eqn:3dgs-alpha-blending}). This mismatch limits Feature-3DGS’s ability to reliably segment objects in 3D, whereas our method addresses this limitation by directly leveraging gradient information.

Another class of methods that segments objects from few visual prompts \cite{cen2023saga, hu2024sagdboundaryenhancedsegment3d}. Our approach is similar to that of SAGD Boundary-Enhanced Segmentation \cite{hu2024sagdboundaryenhancedsegment3d} in that both methods are training free and use multi-view labeling. Similar to SAGD, our approach is training-free and utilizes multi-view labeling. However, while SAGD uses a binary voting system, we assign influence-based voting to each Gaussian, which allows for more accurate segmentation in complex scenes.

Our method is based on the observation that the backpropagated color gradient of each Gaussian indicates its influence on the final rendering (see Section \ref{sec:method}). While inference-time backpropagation is often used for explainability \cite{Selvaraju_2019} and special effects \cite{deepdream} in 2D models, gradient-based methods are generally less suited for clean segmentation. Fortunately, the underlying mathematical formulation of 3D Gaussian Splatting allows for clean segmentation using gradients.

In some applications, especially manipulation tasks, segmenting regions without clear boundaries, such as grasp affordance regions, is challenging, making annotation through language prompting difficult. In such cases, DINO \cite{oquab2024dinov2learningrobustvisual} features can be employed to match and transfer affordance regions. These features are effective for identifying semantically similar regions and have been successfully applied in semantic segmentation\cite{caron2021emergingpropertiesselfsupervisedvision,amir2021deep} and affordance transfer\cite{hadjivelichkov2022oneshottransferaffordanceregions} without any fine-tuning. We utilize DINO-based affordance transfer to map annotated regions from example images to 3D Gaussians. Although the initial transferred affordances may lack alignment, distillation into Gaussian splats improves the result, making it suitable for manipulation tasks.

\section{Methods}
\label{sec:method}
\subsection{3D Segmentation}
3D Gaussian Splatting is a rendering technique that represents scenes using Gaussian distributions as primitives \cite{kerbl3Dgaussians}. Each Gaussian is characterized by its mean position, which defines its center in 3D space, and its covariance, which controls its spatial extent and orientation. This covariance determines the Gaussian’s influence in different directions. Each Gaussian also has associated anisotropic color and opacity.

To render a scene, the Gaussians are depth-ordered relative to the camera, ensuring that nearer Gaussians are rendered on top of those farther away. The 3D Gaussian distributions are then projected onto a 2D plane using the Jacobian of the projection transformation \cite{zwicker2001surface}. This projection determines the size and shape of each Gaussian on the 2D image plane, where the span of each Gaussian represents the region it influences. The opacity of each Gaussian decreases exponentially from its center, resulting in a smooth blending effect. Once the 3D-to-2D transformation is complete, the contributions of all Gaussians are combined using alpha blending to produce the final rendered image (see equation~\ref{eqn:3dgs-alpha-blending}.

{\footnotesize
\begin{lstlisting}[language=Python, caption={Pseudocode of our method}, label={lst:gradient-pseudocode}, frame=single]
def get_3d_mask(gaussians, viewpoints, masks):
    accumulated_grads = [0] * len(gaussians)

    for camera_params, mask_2d in  zip(viewpoints, masks):

        # forward propagation in training mode
        frame = rasterize(gaussians, camera_params) 
        
        # Gradient backpropagation
        accumulated_grads += mask_gradient(gaussians,frame, mask_2d) 
        accumulated_grads -= mask_gradient(gaussians,frame, ~mask_2d)
    mask_3d = accumulated_grads > 0
    return mask_3d
\end{lstlisting}
}

Consider the color $C$ of a pixel at $(x, y)$ in a 3DGS rendering,
\begin{align}
    C &= \sum_{n\leq N} c_n \alpha_n \prod_{m  < n} ( 1 - \alpha_m) \label{eqn:3dgs-alpha-blending} \\ &=\sum_{n\leq N} c_n \alpha_n Tn
\end{align}

Where $N$ is the total number of Gaussians, each indexed by its sorted position, $c_n$ is the color associated with the $n$th Gaussian, $\alpha_n$ is the opacity of the $n$th Gaussian at $(x,y)$ adjusted with exponential falloff, and $Tn=\prod_{m  < n} ( 1 - \alpha_m)$ is the transmittance of $n$th Gaussian at $(x,y)$.

Taking the derivating with respect to color of $k$th Gaussian $c_k$,
\begin{align}
    \frac{ \partial{C}} {\partial{c_k}} = \alpha_k T_k
    \label{eqn:gradient}
\end{align}

This derivative is zero only if either the transmittance $T_k$ or opacity $\alpha_k$ is zero, indicating that the Gaussian does not contribute to the final color. A non-zero gradient indicates that the Gaussian influences the pixel color.

{ \footnotesize
\begin{lstlisting}[language=Python, caption={Pseudocode of 2D-3D affordance transfer}, label={lst:affordance-transfer-pseudocode}, frame=single]
def affordance_transfer(gaussians, labelled_examples):
    """
    labelled_examples are 2D images annotated with affordance regions.
    Unannotated regions are treated as background.
    """
    votes = zeros(m,n) # m in the number of labels, n is the number of gaussians
    example_features = concat([example.features for example in labelled_examples])
    for camera in get_cameras(): # get_cameras is a function that returns an iterator of camera params
        img = rasterize(gaussians, camera)
        features = get_dino_features(img)
        mask = get_init_mask(img)
        # 2d-2d transfer - can be vectorized
        for feature in features:
            mask[feature.region] = kNNSearch(feature, example_features)
        # 2d-3d transfer
        for label in labels: # Including background
            votes[label] += mask_gradient(img, mask == label)
    return argmax(votes, dim=0)
\end{lstlisting}
}
Our approach leverages the gradients of the Gaussians that influence specific 2D segmentation regions. We maintain a buffer, sized according to the number of Gaussians, where gradients are accumulated based on the segmentation masks. Foreground gradients are added to the buffer, while background gradients (from the inverted mask) are subtracted to prevent misclassification (due to influence of background Gaussians near to the surface on foreground region). Once accumulated, the 3D Gaussians with positive gradients in the buffer are classified as part of the segmented region. The pseudocode for this process is provided in Listing \ref{lst:gradient-pseudocode}. %

\subsection{2D-3D affordance transfer}
We first perform a 2D-2D affordance transfer from example images to rendered images. The transferred images serve as segmentation masks. Next, we apply gradient-based voting to the transferred images.

We annotate different affordance regions in example images and extract feature vectors from the output feature map of DINO\cite{oquab2024dinov2learningrobustvisual} corresponding to those examples. With feature vectors for each affordance region (including the background), we classify each patch in the rendering using kNN with cosine similarity, where the example feature vectors act as the training set. Each patch is then assigned a label.

Finally, we apply the voting algorithm (shown in Listing \ref{lst:affordance-transfer-pseudocode}). Even if the input affordance map contains inaccurate patches, the error is compensated after voting across multiple frames.

\begin{figure}
    \centering
    \begin{tabular}{ccc}
        \adjustbox{valign=c}{\includegraphics[width=0.15\textwidth]{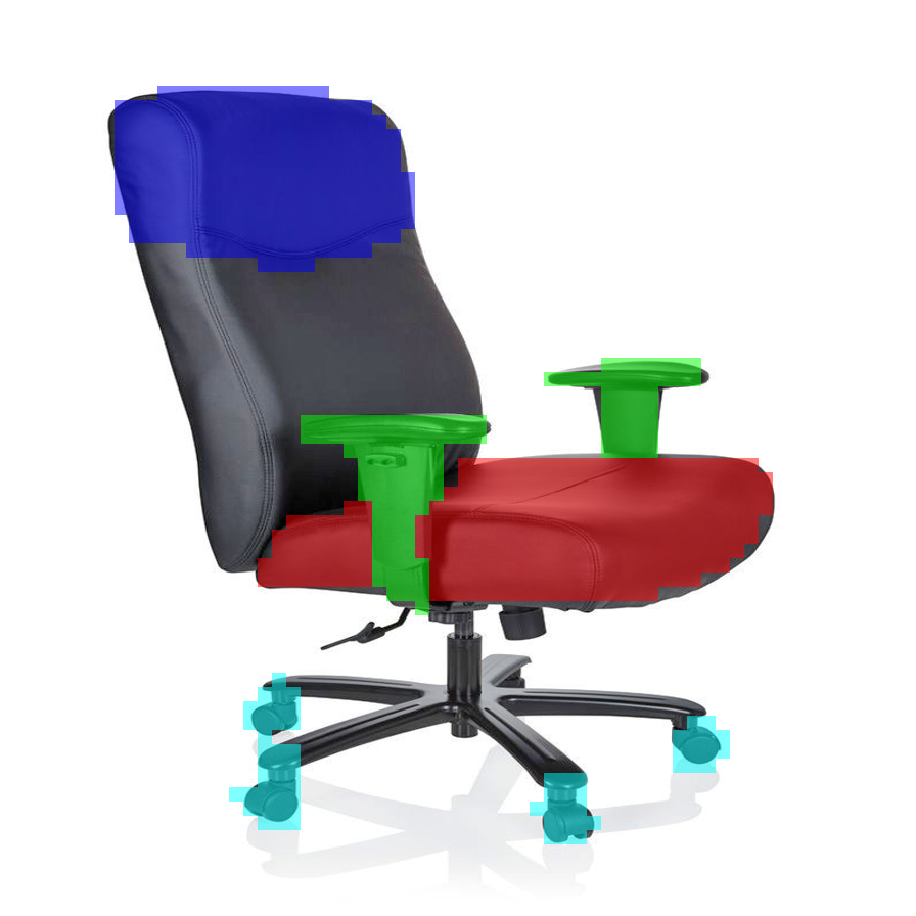}} &
        \adjustbox{valign=c}{\includegraphics[width=0.125\textwidth]{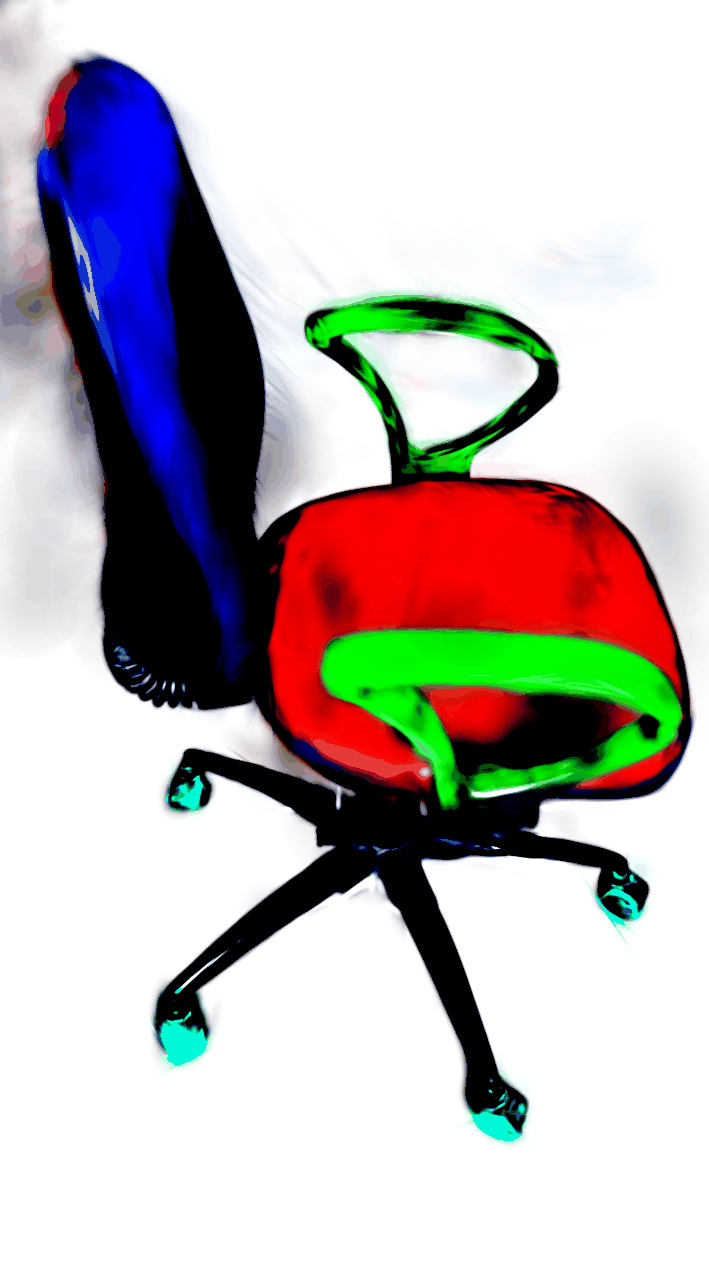}} &\\
        (a) & (b)\\
        \adjustbox{valign=c}{\includegraphics[width=0.125\textwidth]{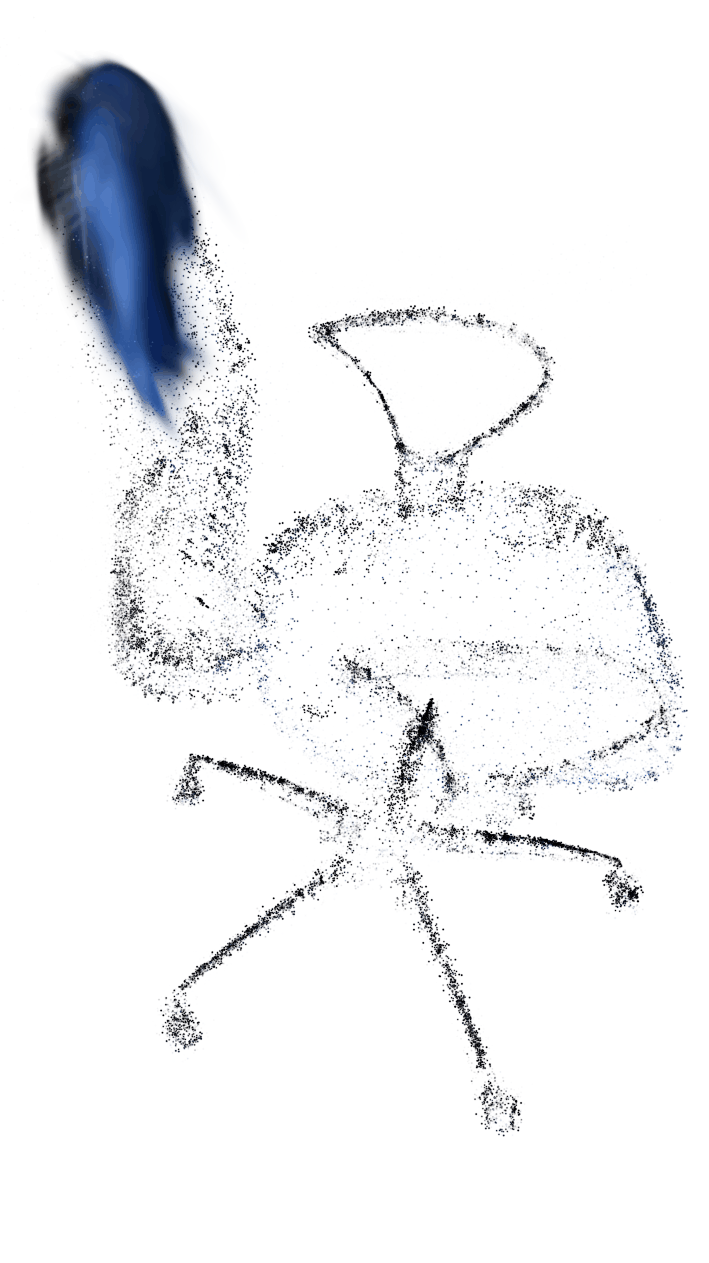}} &
        \adjustbox{valign=c}{\includegraphics[width=0.125\textwidth]{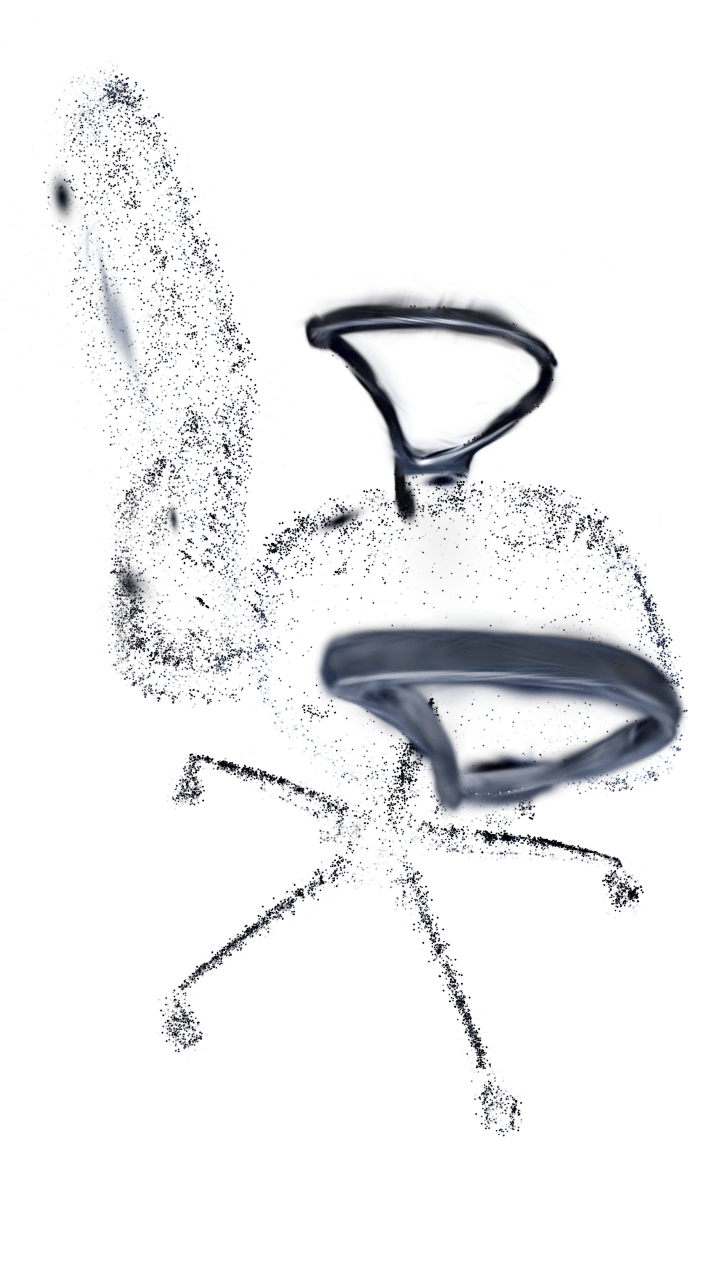}} &\\
        (c) & (d)\\
        \adjustbox{valign=c}{\includegraphics[width=0.125\textwidth]{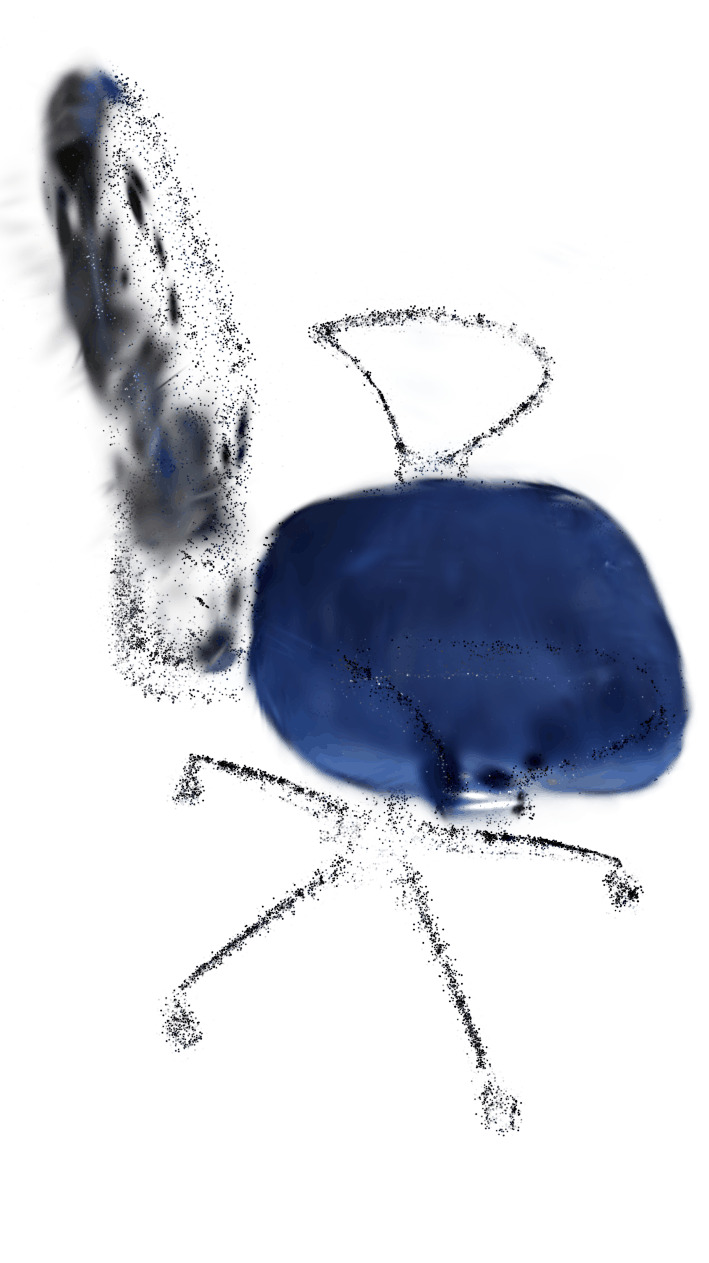}} &
        \adjustbox{valign=c}{\includegraphics[width=0.125\textwidth]{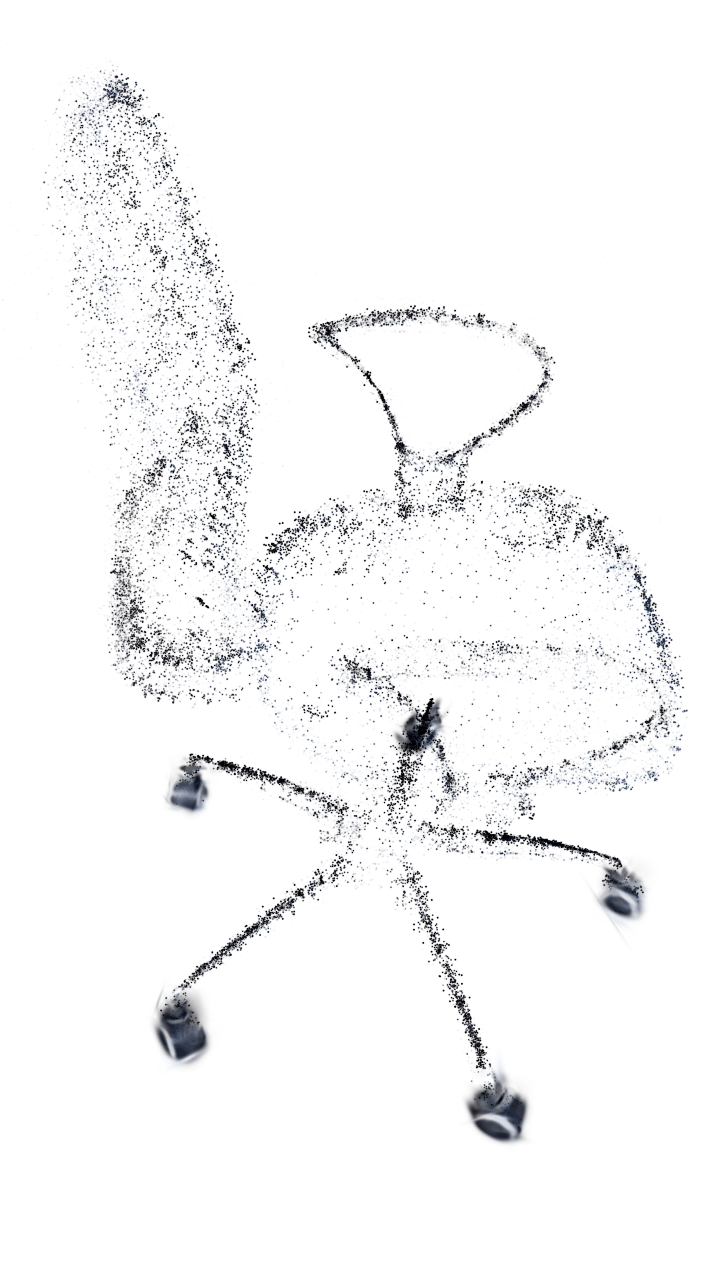}} &\\
        (e) & (f)\\
        
    \end{tabular}
    \caption{\textbf{(a)} Example image with annotated affordances. \textbf{(b)} Segmentation map showing transferred affordances in a 3D scene. \textbf{(c)-(f)} Visualizations of different parts, with the remaining Gaussians represented as a point cloud for clarity, allowing better understanding of regions in relation to the whole scene.}

    \label{fig:seg-comparison-with-feature-3dgs}
\end{figure}

\section{Experiments}

\subsection{Gradient Calculation}

\begin{figure*}[h!]
    \centering
    \begin{tabular}{ccc}
        \toprule
        \textbf{Input Mask} & \textbf{Extraction} & \textbf{Deletion} \\
        \midrule
        \adjustbox{valign=c}{\includegraphics[width=0.3\textwidth]{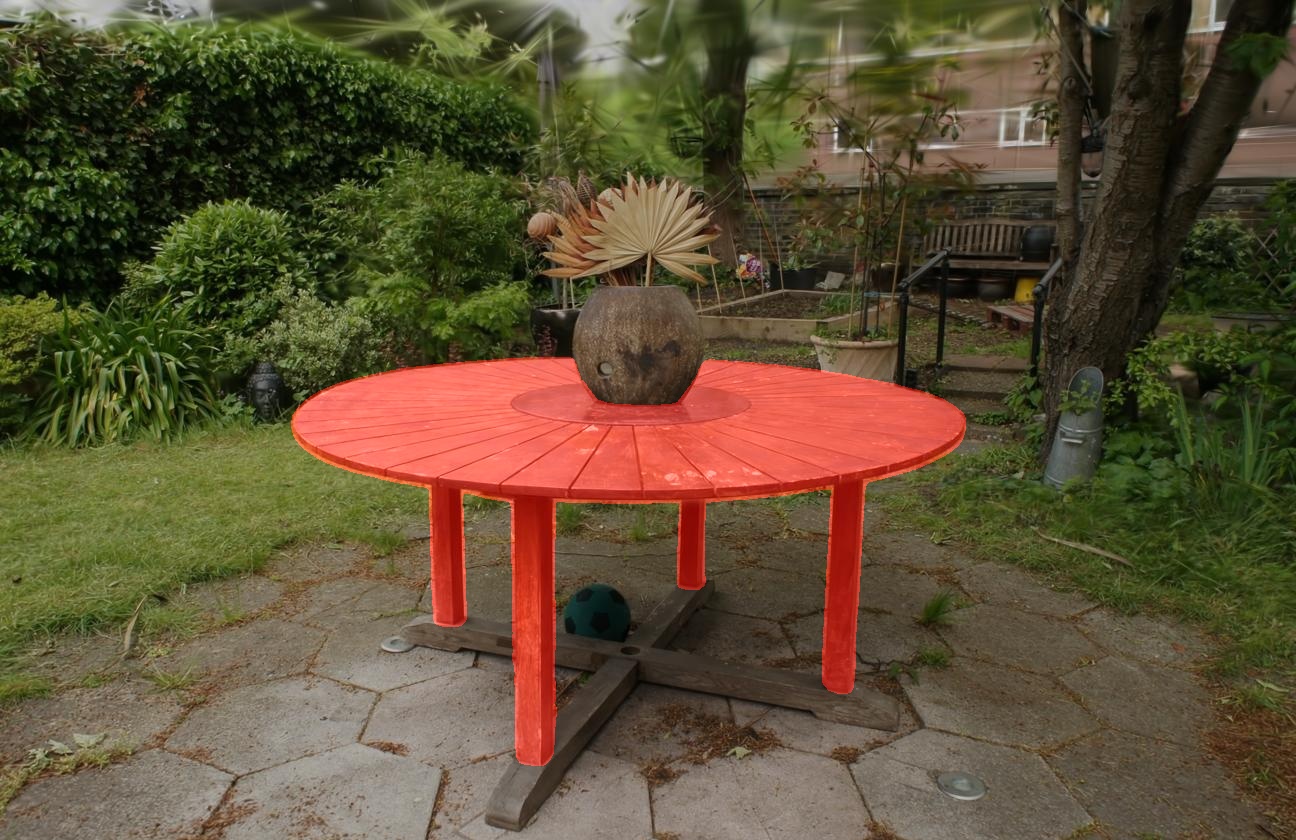}} &
        \adjustbox{valign=c}{\includegraphics[width=0.3\textwidth]{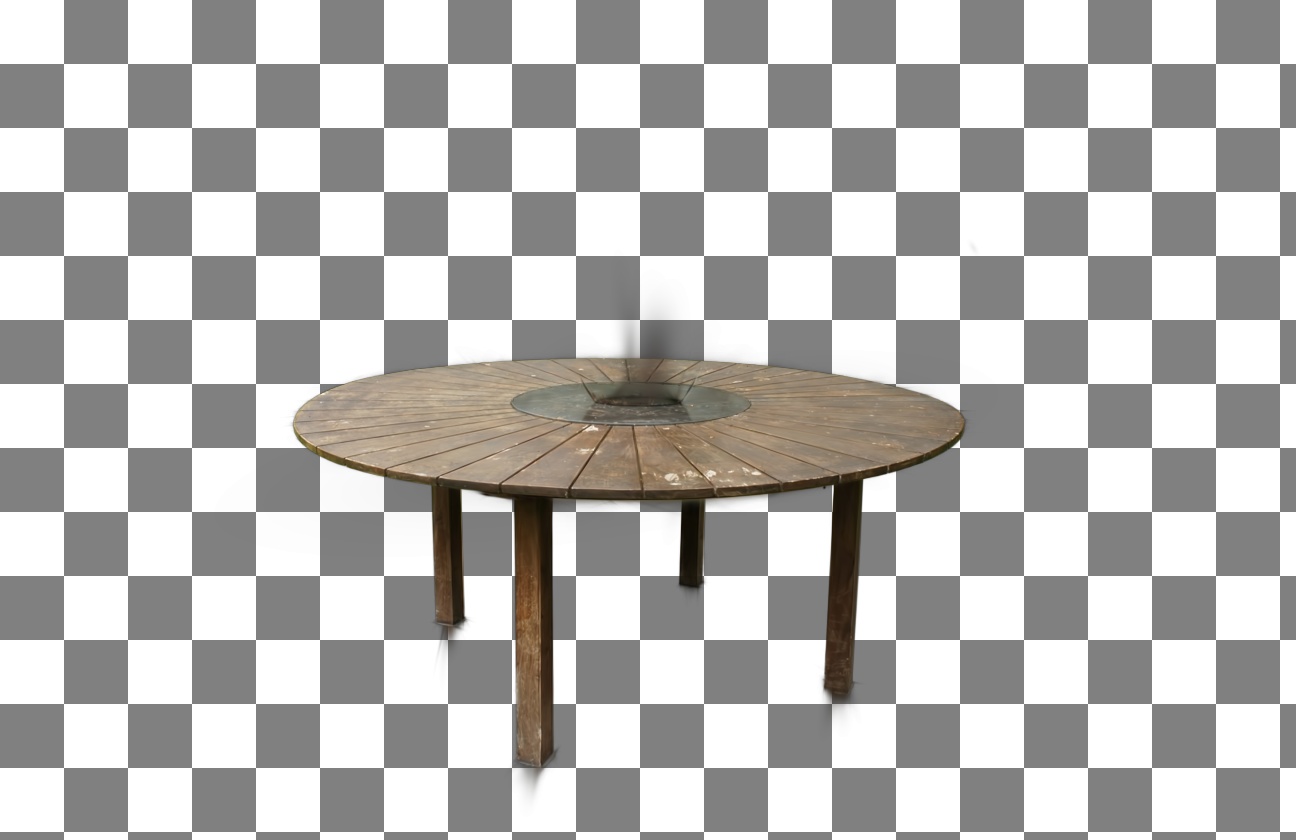}} &
        \adjustbox{valign=c}{\includegraphics[width=0.3\textwidth]{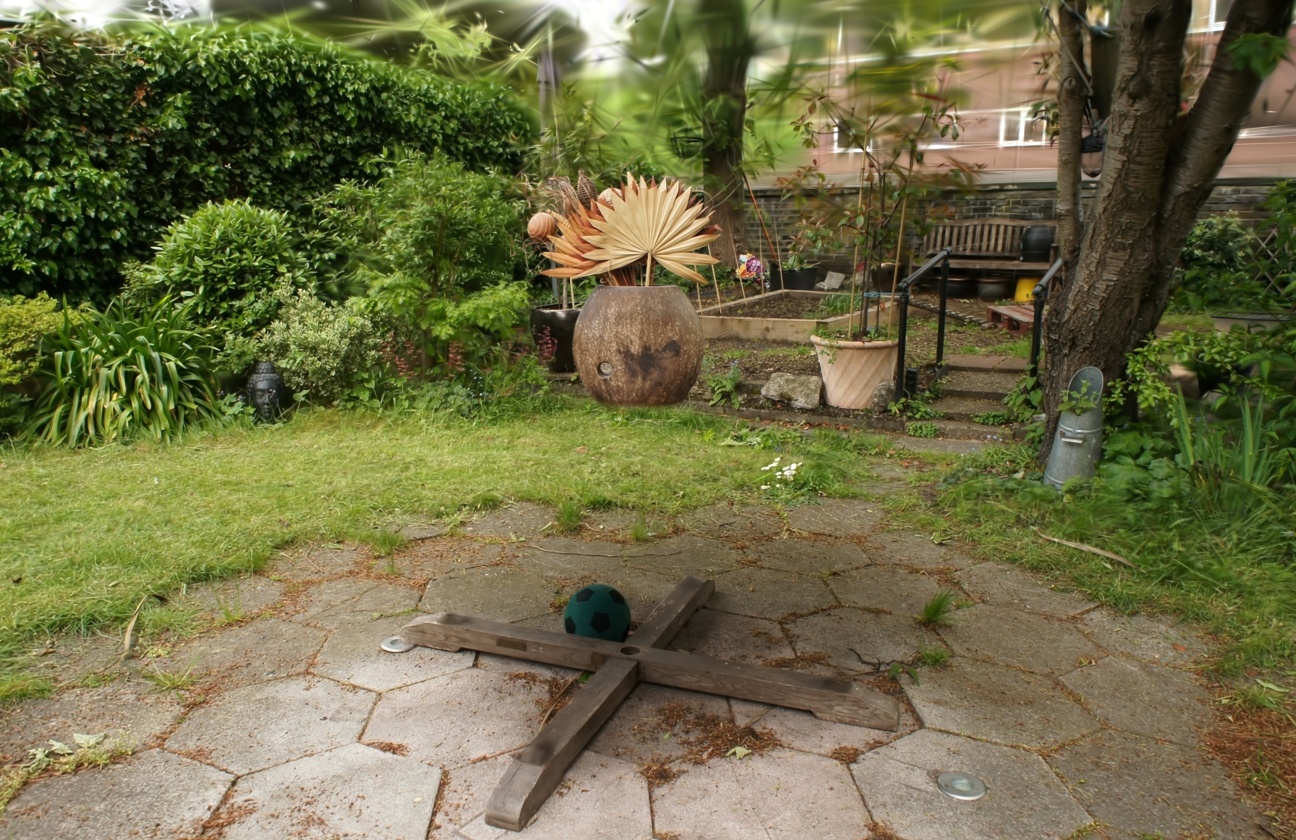}} \\
        \adjustbox{valign=c}{\includegraphics[width=0.3\textwidth]{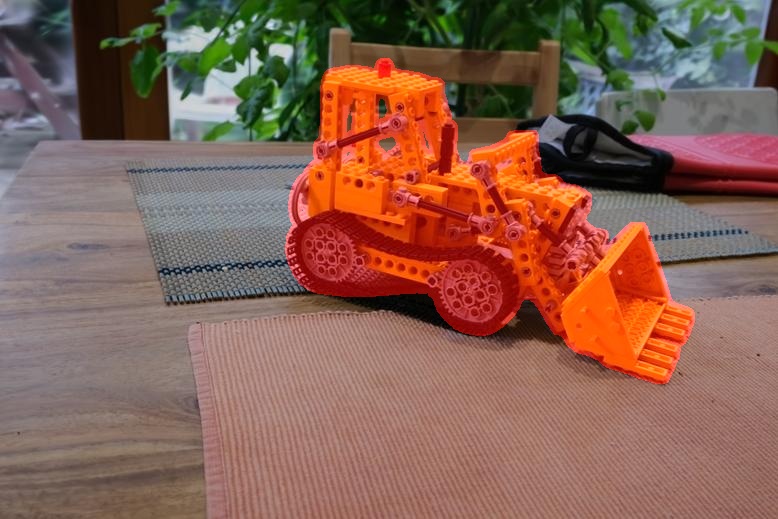}} &
        \adjustbox{valign=c}{\includegraphics[width=0.3\textwidth]{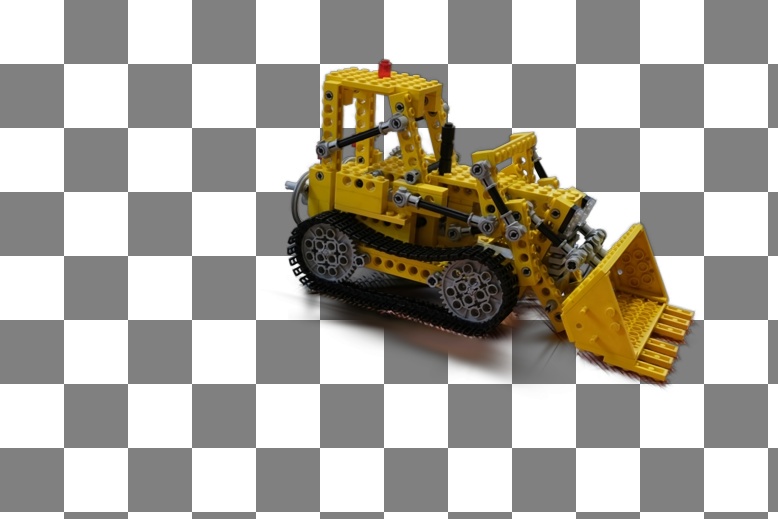}} &
        \adjustbox{valign=c}{\includegraphics[width=0.3\textwidth]{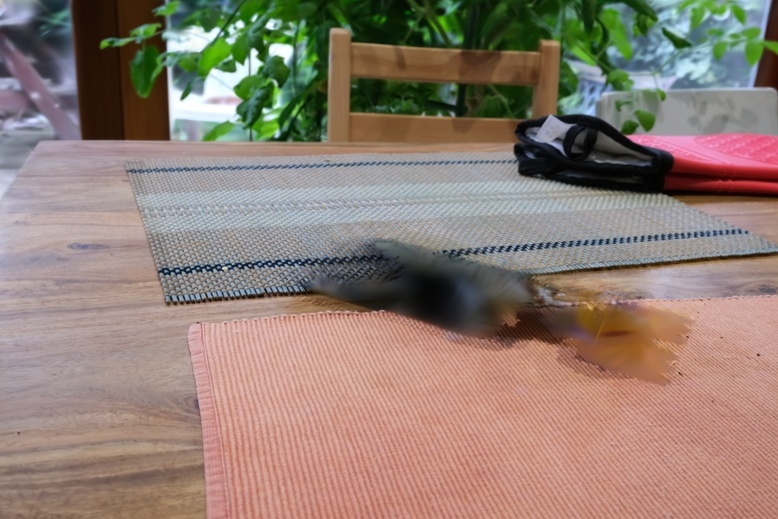}} \\
        \adjustbox{valign=c}{\includegraphics[width=0.3\textwidth]{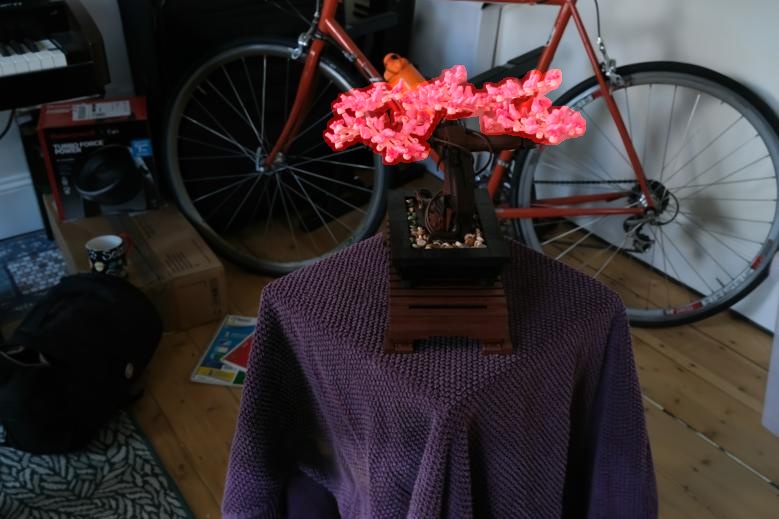}} &
        \adjustbox{valign=c}{\includegraphics[width=0.3\textwidth]{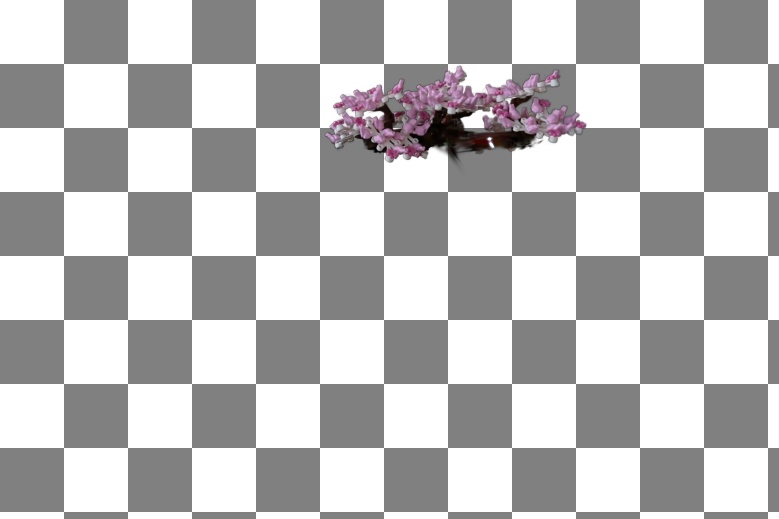}} &
        \adjustbox{valign=c}{\includegraphics[width=0.3\textwidth]{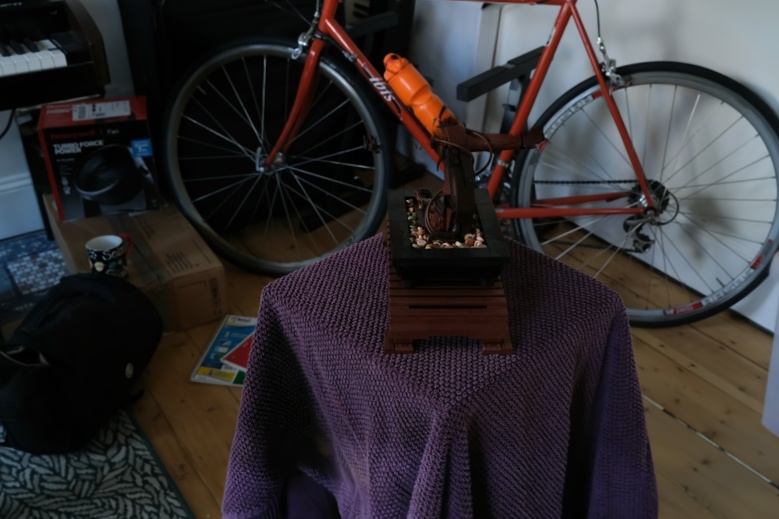}} \\
        \bottomrule
    \end{tabular}
    \caption{The first column shows a rendered frame along with its corresponding input mask. In the second column, we present the results after extracting the 3D Gaussians that align with the generated 3D mask. The third column illustrates the rendered output from the remaining Gaussians. Notably, the background is visible instead of a blank space, as the segmentation occurs directly within the 3D space. For enhanced clarity, zooming in is recommended.}
    \label{fig:seg-quantitative-examples}
\end{figure*}
We utilized gsplat \cite{ye2023mathematical} as our 3D Gaussian rasterizer, where the color of each Gaussian was represented by spherical harmonics (SH) coefficients. Since the gradients were independent of color values, as demonstrated in Equation~\ref{eqn:gradient}, we treated the DC components of the SH coefficients as analogous to color and calculated gradients with respect to these DC components.

Once the image was rendered, we multiplied the output by the mask element-wise, computed the mean, and backpropagated the gradients to the input color (represented by SH coefficients). All experiments were conducted on an NVIDIA A6000 GPU.

\subsection{3D Segmentation}
To identify the segmentation mask, we used YOLO-world \cite{Cheng2024YOLOWorld} for the initial bounding box detection, followed by SAM 2 \cite{ravi2024sam2} to estimate and track masks throughout the rendered frames.

Some of the qualitative results were shown in Figure \ref{fig:seg-quantitative-examples}. We provided additional results on the project page.

For the quantitative comparison, we employed the Mip-NeRF 360 dataset. To ensure a fair, apple-to-apple comparison, we compared our method against two baseline voting methods.

\textbf{Baseline 1:} This method assigns a vote if the projected Gaussian falls within the input 2D mask. A limitation of this approach is that it still votes for Gaussians that are occluded.

\textbf{Baseline 2:} This method uses gradient magnitude to decide which Gaussians receive votes, but the voting magnitude is constant for all Gaussians. A drawback is that the foreground need not be fully opaque, even though it appears so. As a result, some background Gaussians may still receive votes.

Since we did not have access to the actual 3D volume, we repurposed the 2D mIoU metric for evaluation. We uniformly sampled 10\% of the masks for segmentation purposes, while using the remaining 90\% for evaluation.

To produce the 2D mask from the estimated 3D mask, we assigned black to the background and white to the foreground. Then we thresholded the grayscale rendered output to obtain the estimated mask. We did not use segmented 3D region to produce 2D masks because the ground truth 2D masks were often partially occluded, while the 2D masks from extracted regions were unoccluded.

The results are shown in Table \ref{tab:3d-segmentation-quantitative}. We report the mean Intersection over Union (mIoU) between the estimated masks and the ground truth masks. We chose not to report pixelwise accuracy because, due to the typically larger background regions, it tends to yield a high value that does not accurately reflect the quality of the segmentation.

\begin{table}[ht!]
    \caption{Table showing the comparison of mIoU between our method and baselines on scenes where we were able to track object using SAM 2. We used 10\% of the images for segmentation purposes, while the remaining 90\% were used for evaluation.}
    \label{tab:3d-segmentation-quantitative}
    \centering
    \begin{tabular}{lccc}
    \toprule
       \textbf{Scene} & \textbf{Baseline 1} & \textbf{Baseline 2} & \textbf{Ours} \\
       \midrule
       Bicycle & 52.03 & 55.55 & \textbf{59.46} \\
       Bonsai & 77.48 & 78.67 & \textbf{81.44} \\
       Garden & 66.17 & 81.78 & \textbf{95.23} \\
       Kitchen & 80.70 & 82.79 & \textbf{92.89} \\
       Room & 93.79 & 89.60 & \textbf{94.57} \\
       \midrule
       Mean & 74.03 & 77.68 & \textbf{84.72} \\
       \bottomrule
    \end{tabular}
\end{table}

As seen in Table \ref{tab:3d-segmentation-quantitative}, our method outperforms both baselines across all scenes. In particular, the Bicycle scene posed a challenge due to the presence of two objects (a bicycle and a bench), both of which contain many thin structures and holes. These structures were difficult to segment accurately because the ground truth masks covered the holes, leading to a reduction in mIoU for this scene.

\subsection{Affordance Transfer}

\begin{figure}
    \centering
    \includegraphics[width=0.8\linewidth]{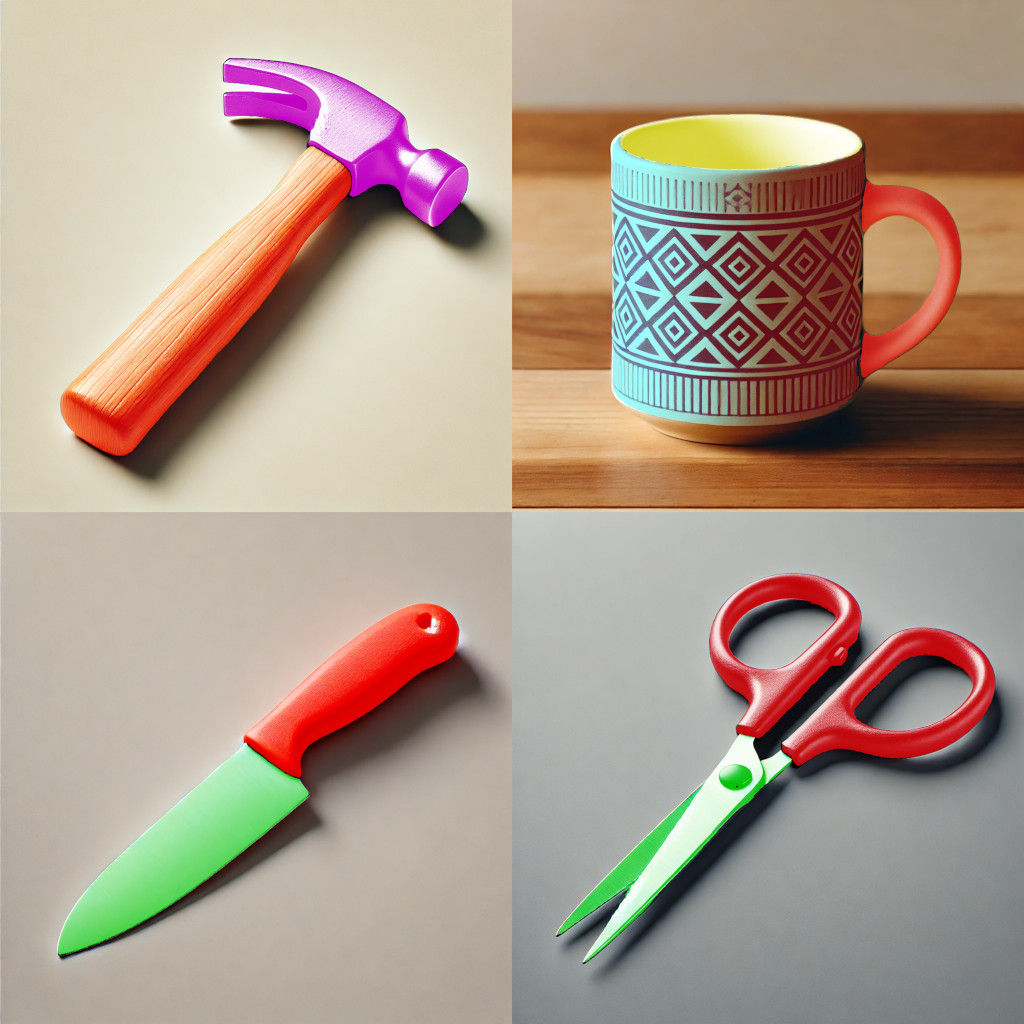}
    \caption{Examples of images used for affordance transfer with annotations. Note that these instances, though belonging to the same class, are different from the objects in the target frames.}
    \label{fig:examples-affordance-transfer}
\end{figure}

For each scene, we annotated a few examples with affordance categories using LabelMe \cite{Wada_Labelme_Image_Polygonal}. The images were generated using DALL-E and, although they belong to the same category, they are different from the objects in the 3D scene. Figures~\ref{fig:examples-affordance-transfer} shows few of the example images along with labels. We rendered each image and transferred the affordance regions from the example images to the rendering. The voting algorithm, using masked gradients, was applied as described in Listing \ref{lst:affordance-transfer-pseudocode}.

For quantitative evaluation, we used the RGB-D Part Affordance Dataset \cite{Myers:ICRA15}. We constructed the Gaussian splats using the provided cluttered scenes and then calculated the mIoU after the affordance transfer. The quantitative results are shown in Table \ref{tab:affordance-quantitative}.

Since the affordance regions were estimated using 2D annotated examples rather than the ground truth, we used all manually annotated images (provided by the authors of \cite{Myers:ICRA15}) for evaluation. Some images in the original dataset were automatically annotated, which resulted in significant differences in the ground truth. Therefore, we relied solely on manually annotated frames for a more consistent evaluation.

\begin{figure}%
    \begin{subfigure}[b]{0.24\textwidth}\includegraphics[width=\textwidth]{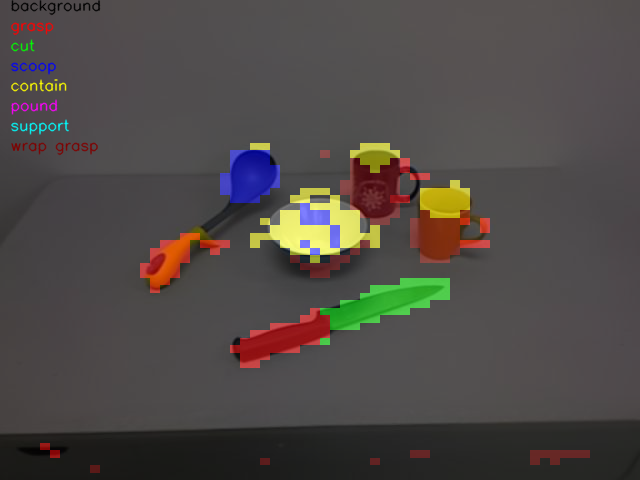}
        \caption{2D-2D Affordance Transfer}
    \end{subfigure} %
    \begin{subfigure}[b]{0.24\textwidth}\includegraphics[width=\textwidth]{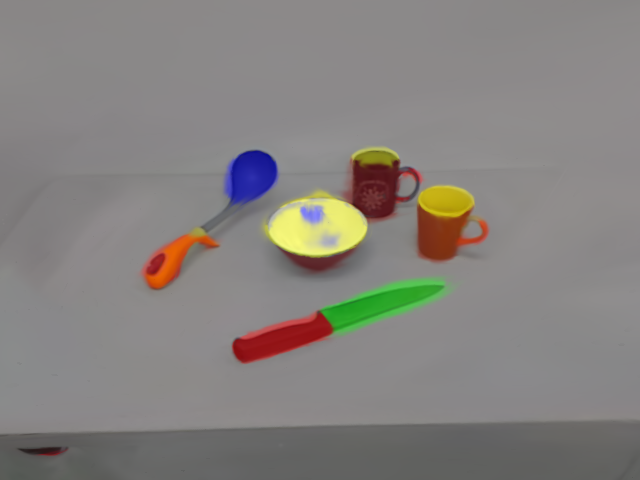}
        \caption{2D-3D Affordance Transfer}
    \end{subfigure} %
    \caption{Results of 2D-2D and 2D-3D affordance transfer. The labels generated during the 2D-2D affordance transfer serve as input to the 2D-3D affordance transfer. Despite not having perfectly aligned labels, voting over multiple frames makes 2D-3D affordance transfer more precise.}
    \label{fig:2d-2d_vs_2d-3d}
\end{figure}

\begin{table}[htbp!]
    \caption{Table showing comparison of 2D-2D and 2D-3D affordance transfer.}
    \label{tab:affordance-quantitative}
    \centering
    \begin{tabular}{lcccc}
        \toprule
        & \multicolumn{2}{c}{\textbf{mIoU}} & \multicolumn{2}{c}{\textbf{Recall}} \\
        \cmidrule(r){2-3} \cmidrule(l){4-5}
        \textbf{Scene} & \textbf{2D-2D} & \textbf{2D-3D} & \textbf{2D-2D} & \textbf{2D-3D} \\
       \midrule
       1 & 40.09  & \textbf{47.87} & 61.37 & \textbf{67.77}\\
       2 & 45.92   & \textbf{55.63} & 69.80 & \textbf{81.07}\\
       3 & 49.48   & \textbf{60.50} & 73.46 & \textbf{86.95}\\
       \midrule
        Mean & 45.16 & \textbf{54.67} & 68.21  & \textbf{78.60} \\
       \bottomrule
    \end{tabular}
\end{table}

Since the affordance transfer is conducted patch-wise, objects with narrow regions may result in nearby background regions receiving votes. To address this, we introduced recall—true positive pixels divided by the total number of pixels in the ground truth mask—to discard pixels that fall outside the ground truth mask, providing a clearer interpretation of the results.

Our hypothesis was that less accurate 2D-2D transfer could still lead to accurate 2D-3D transfer due to the large number of pooled inputs in the voting process. The results in Table~\ref{tab:affordance-quantitative} and Figure~\ref{fig:2d-2d_vs_2d-3d} confirm this hypothesis, demonstrating that pooling from multiple frames improves the accuracy of the 2D-to-3D affordance transfer.

\subsection{Pruning Gaussians with Inference-Time Backpropagation}
For pruning, we considered the entire image as a mask and voted for each Gaussian based on its gradient magnitude. At the end of the process, we retained only the Gaussians with non-zero votes.

We show the results of applying our method to the Mip-NeRF 360 dataset in Table \ref{tab:pruning}.

\begin{table}[]
    \centering
    \caption{Each scene in the Mip-NeRF dataset and the percentage of Gaussians pruned.}
    \label{tab:pruning}
    \begin{tabular}{lr}
        \toprule
         \textbf{Scence} & \textbf{Gaussians Removed (\%)} \\
         \midrule
         bicycle & 3.06 \\
         bonsai & 11.92 \\
         garden & 0.65 \\
         kitchen & 3.97 \\
         room & 21.49 \\
         counter & 7.06 \\
         stump & 1.40\\
         \midrule
         Mean & 7.08 \\
         
         \bottomrule
    \end{tabular}
\end{table}

The size of each scene was capped at 1,000,000 Gaussians, and the training process followed the MCMC strategy from~\cite{kheradmand20243dgaussiansplattingmarkov}. We verified the maximum absolute pixel error across all viewpoints after pruning, which remained at zero, indicating that our method efficiently pruned Gaussians without compromising accuracy.

\section{Conclusion}

We presented a method that utilizes inference-time backpropagation to segment 3D Gaussian splats and few shot affordacne transfer. With a minor modification, our approach was able to prune trained Gaussian splatting models without any loss in accuracy.

However, this method does have limitations, primarily its dependence on accurate 2D masks and chosen viewpoints. Moreover, the gradient calculations across multiple frames result in slower performance compared to feature-field-based methods.

Despite these drawbacks, the method’s high segmentation quality makes it highly suitable for downstream applications that require one-time or soft-real-time segmentation, such as digital twins, augmented reality, and asset generation.

\bibliographystyle{abbrv}
\bibliography{references}

\end{document}